\documentclass{svproc}
\usepackage{graphicx}%
\usepackage{multirow}%
\usepackage{amsmath,amssymb,amsfonts}%
\usepackage{bbm}%
\usepackage{mathabx,mathrsfs}%
\usepackage[title]{appendix}%
\usepackage{xcolor}%
\usepackage{textcomp}%
\usepackage{manyfoot}%
\usepackage{booktabs}%
\usepackage{algorithm}%
\usepackage{algorithmicx}%
\usepackage{algpseudocode}%
\usepackage{pifont}%
\usepackage{stmaryrd}%
\usepackage{listings}%
\usepackage{url}

\begin{document}
\mainmatter              % start of a contribution
\title{Optimal Transport for Network Comparison: \\
A Unified Review with New Spectral Bounds and Machine Learning Applications}
\titlerunning{Optimal Transport for Network Comparison}  % abbreviated title (for running head)
\author{James Hyun\inst{1} \and Fran\c{c}ois G. Meyer\inst{1}}
% Author3Name Author3Surname \and Author4Name Author4Surname \and Author5Name Author5Surname \and Author6Name Author6Surname}
%
\authorrunning{James Hyun and Fran\c{c}ois G. Meyer} % abbreviated author list (for running head)

\institute{University of Colorado, Boulder, CO 80309,\\
\email{sehy1420@colorado.edu}, \email{fmeyer@colorado.edu}\\ % WWW home page:
% \texttt{http://users.com/author1.html}
% \and
% Institution2,
% address,\\
% state, country}
United States
}

\maketitle              % typeset the title of the contribution

\begin{abstract}
Network comparison using optimal transport is a growing area of research in network science. Unlike standard graph metrics, optimal transport computes both network dissimilarity and a transport plan that explains how one graph morphs into another. In this paper, we review how optimal transport compares undirected, unweighted simple graphs using three primary distances: the Wasserstein, Gromov-Wasserstein, and Bures-Wasserstein distances. We examine the closed form of the Wasserstein distance in one dimension via node feature probability distributions, and show how the transport plans of the Wasserstein and Gromov-Wasserstein distances visualize how mass is shifted to transform one network into another. Beyond reviewing existing transport-based approaches, we establish new spectral lower and upper bounds for the Bures-Wasserstein distance and characterize the tightness of the lower bound under eigenbasis perturbations. Finally, we evaluate these distances using a synthetic network dataset for clustering and a real-world temporal network.

\end{abstract}
\section{Introduction}\label{sec1}

Networks, or graphs, are fundamental data for complex systems with interacting components, arising in domains ranging from neuroscience and social science to biology and communication systems. Quantifying the distance between graphs has become a key focus in data science, providing a way to measure the dissimilarity of complex relationships and interactions. However, graphs are hard to analyze because they do not live in Euclidean space. Motivated by the Gromov-Hausdorff distance in shape analysis (\cite{memoli2007use}, \cite{memoli2008gromov}) -- which measures how far two compact metric spaces are from being isometric -- several studies have sought to adapt this metric to define the network distance (\cite{chowdhury2018metric}, \cite{chowdhury2023distances}). Although it is mathematically well-defined pseudo-metric in the set of all graphs, it is NP-hard to compute.

To avoid NP-hard computation, vector- and matrix-based distances have become popular to compare graphs by embedding graphs into vectors or matrices. Although these distances are easy to compute, they depend on graph size or require the graph alignment (\textit{e.g.,} DeltaCon). Conversely, the graph edit distance is independent of graph size but is NP-hard to compute. We refer the reader to \cite{wills2020metrics} for more details on graph distances. 

To avoid these bottlenecks, optimal transport has emerged as a powerful field for network comparison. Because optimal transport compares probability measures, applying it to network-valued data requires embeddings that transforms a network into a probability distribution. Our work focuses on how different network embeddings can be equipped to optimal transport-based distances to induce fundamentally different notions of network similarity. The remainder of the paper is organized as follows: 
\begin{itemize}
    \item Section 2 reviews the necessary foundation of graph theory and optimal transport.

    \item Section 3 discusses the Wasserstein, Gromov-Wasserstein, and Bures-Wasserstein distances to compare networks across different embedding spaces. We also present spectral bounds for the Bures-Wasserstein distance between graph signal distributions, alongside a perturbation bound characterizing the tightness of the lower bound. 

    \item Section 4 applies these distances to a circulant graph, network clustering, and a temporal network.

    \item Section 5 concludes the paper by summarizing each transport distance for network comparison and outlines directions for future research.
\end{itemize}
In short, we provide a unified optimal transport perspective for network comparison. We hope that this paper provides practical guidance on the use of optimal transport-based distances in network analysis.

\section{Optimal Transport in Networks}\label{sec2}

\subsection{Graph Theory}\label{subsec1}

An \textbf{undirected unweighted graph} $G=(V,E)$ consists of a node set $V$ and an edge set $E$. If two nodes $i$ and $j$ are connected by an edge, we write $(i,j)\in E$ or $i\sim j$. A \textbf{digraph} is a graph whose edges have a specific orientation, indicating a one-way relationship. A \textbf{connected graph} is an undirected graph where a path exists between every pair of nodes. A \textbf{simple graph} is an undirected unweighted graph that contains no self-loops and no more than one edge between two nodes.

An \textbf{adjacency matrix} $A=[A_{ij}]\in \mathbb{R}^{|V|\times|V|}$ is a symmetric matrix with entries $A_{ij}=1$ if $(i,j)\in E$ and $A_{ij}=0$ otherwise. We write the \textbf{degree matrix} $D$, and the \textbf{(unnormalized) Laplacian matrix} is a symmetric positive semi-definite matrix, defined as $L=D-A$. The multiplicity of the zero eigenvalue of $L$ is equal to the number of connected components in an undirected graph (\cite{brouwer2011spectra}). 

Throughout, we refer to a graph as a simple graph and write $G=(V,E)$ and $G'=(V',E')$ with $|V|=n$, $|V'|=m$, and $n\neq m$, unless otherwise stated. The Laplacian matrices of $G$ and $G'$ are $L$ and $L'$, respectively, and their spectral decompositions are $L=U\Lambda U^T$ and $L'=U'\Lambda'(U')^T$. The eigenvalues $\{\lambda_i\}_{i=1}^n$ of $L$ and $\{\lambda_j'\}_{j=1}^m$ of $L'$ are in ascending order.

\subsection{Optimal Transport}\label{subsec2}

Optimal transport is to compute the minimal cost of moving masses in one probability space $(X,\mu)$ to another probability space $(Y,\nu)$. The probability measure $\pi$ in the product space $X\times Y$ whose marginals are $\mu$ and $\nu$ is called a \textbf{transport plan}. If $\mu=\sum_{i=1}^n a_i\delta_{x_i}$ and $\nu=\sum_{j=1}^m b_j\delta_{y_j}$ are probability measures in $\mathbb{R}^d$, then $\pi=[\pi_{ij}]\in \mathbb{R}_+^{n\times m}$, and $\pi_{ij}$ represents the mass transported from bin $i$ to bin $j$. We denote by $\Pi(\mu,\nu)$ the set of all transport plans.

Let $C_{ij}=c(x_i,y_j)$ be the transport cost from bin $i$ to bin $j$ for some measurable cost function $c:\mathbb{R}^d\times \mathbb{R}^d\rightarrow \mathbb{R}_+$. The \textbf{Kantorovich problem} minimizes the Frobenius inner product between $C$ and $\pi$ over all transport plans in $\Pi(\mu,\nu)$.
\begin{equation}
    \text{minimize}\:\sum_{i=1}^n\sum_{j=1}^m C_{ij}\pi_{ij},\quad\text{s.t.}\quad\begin{cases}
        \sum_{j=1}^m \pi_{ij}=a_i,\quad &\forall i\in \llbracket n\rrbracket,\\
        \sum_{i=1}^n\pi_{ij}=b_j,\quad &\forall j\in \llbracket m\rrbracket,\\
        \pi_{ij}\geq 0, &\forall i\in \llbracket n\rrbracket,\forall j\in \llbracket m\rrbracket.
    \end{cases}\tag{KP}
\end{equation}
If we require moving mass from one location to a single target without splitting it, then (KP) becomes the \textbf{Monge problem}.
\begin{align}
    \text{minimize}\:\sum_{i=1}^nc(x_i,T(x_i))a_i,\quad\text{s.t.}\quad T_\#\mu=\nu.\tag{MP}
\end{align}
Problem (MP) seeks the optimal transport map $T:X\rightarrow Y$ such that $\nu$ is the \textbf{pushforward measure} of $\mu$, and such the map is called the \textbf{Monge map}.

For probability measures on $\mathbb{R}$ and a strictly convex cost function depending on $|x-y|$ (\textit{e.g.,} $c(x,y)=|x-y|^2/2$), the Monge map $T$ is uniquely given by the closed form $T=G^{-1}\circ F$, where $F$ and $G$ are the cumulative distribution functions of $\mu$ and $\nu$, respectively. Moreover, $T$ is monotonic and $\pi=(\text{Id}\times T)_\#\mu$, where $\text{Id}$ is the identity map on $\mathbb{R}^d$. For more properties of the Monge map, we refer the reader to \cite{villani2021topics} and \cite{ambrosio2012user}.

The exact time complexity of solving (KP) when $n=m$ is $\mathcal{O}(n^3\log n)$ (\cite{pele2009fast}). However, the Sinkhorn algorithm (\cite{cuturi2013sinkhorn}, \cite{peyre2019computational}) approximates (KP) by adding an entropic regularization term $-\epsilon H(\pi)=\epsilon\sum_{i=1}^n\sum_{j=1}^m \pi_{ij}(\log\pi_{ij}-1)$. The Gibbs kernel $K_{ij}=\exp\{-C_{ij}/\epsilon\}$ and the KKT condition in convex analysis reduce the number of $nm$ unknowns in (KP) by finding $u=\{u_i\}_{i=1}^n$ and $v=\{v_j\}_{j=1}^m$. The algorithm uses fixed point iterations $u_{k+1}=\mu/Kv_k$ and $v_{k+1}=\nu/K^Tu_{k+1}$ and reduces the cubic time complexity. We refer the reader to Table 1 in \cite{luo2023improved} for the improvement of algorithms for solving optimal transport problems.

\section{Transport Distances Between Networks}\label{sec3}

\subsection{Discrete Network Embedding}\label{subsec4}

\subsubsection{Empirical measure}\label{subsubsec1}

One particular case of (KP) widely used in practice is when $C$ is the distance matrix where each entry is the distance between the support points of the empirical measures $\mu$ and $\nu$. This gives the famous probability metric, called the Wasserstein distance.

\begin{definition}[Wasserstein Distance, \cite{peyre2019computational}]
Let $\mu=\sum_{i=1}^n a_i\delta_{x_i}$ and $\nu=\sum_{j=1}^m b_j\delta_{y_j}$ be probability measures in $(\mathbb{R}^d,\|\cdot\|)$. For $1\leq p<\infty$, the $p$-Wasserstein distance between $\mu$ and $\nu$ is defined as
\begin{align}
    W_p(\mu,\nu)=\bigg(\min_{\pi\in \Pi(\mu,\nu)}\sum_{i=1}^n\sum_{j=1}^m \|x_i-y_j\|^p\pi_{ij}\bigg)^\frac{1}{p}.
\end{align}
\end{definition}
To apply $W_p$ for network comparison, we embed two graphs $G$ and $G'$ into the node feature distributions $\mu_G=\frac{1}{n}\sum_{i=1}^n\delta_{x_i}$ and $\mu_{G'}=\frac{1}{m}\sum_{j=1}^m\delta_{y_j}$. Because $W_p$ compares two graphs as the probability distributions, unequal graph sizes ($n\neq m$) present no conceptual difficulty. Throughout the paper, we study $W_1$ because it bounds $W_p, p\neq 1$ by \textit{H\"older's inequality} under the assumption of bounded supports (\cite{villani2021topics}). 

While computing $W_p$ in high dimensions may be infeasible to compute, if $\mu_G$ and $\mu_{G'}$ are supported on $(\mathbb{R},|\cdot|)$, then $W_1(\mu_G,\mu_{G'})$ simplifies to the $L^1$-metric between their cumulative distribution functions (\cite{peyre2019computational}).
\begin{align}
    W_1(\mu_G,\mu_{G'})=\int_\mathbb{R}\bigg|\frac{1}{n}\sum_{i=1}^n \mathbbm{1}_{[x_i,\infty)}(t)-\frac{1}{m}\sum_{j=1}^m \mathbbm{1}_{[y_j,\infty)}(t)\bigg|dt.
\end{align}
The node features $\{x_i\}_{i=1}^n$ and $\{y_j\}_{j=1}^m$ are sorted in ascending order, and (2) is computable in $\mathcal{O}(n\log n+m\log m)$ time. We note that $W_p(\mu_G,\mu_{G'})$ for $p>1$ admits the closed form in terms of the generalized inverse of $\mu_G$ and $\mu_{G'}$.

\subsubsection{Node feature}\label{subsubsec2}

Any node feature, such as degree, centrality, clustering coefficient, graphlet count, PageRank, etc., can be used in (2) as long as the feature does not blow up under graph perturbation. \textbf{Fig. 1} shows harmonic centrality used as a feature in the stochastic block model with 100 nodes and two communities of size 50.

\begin{figure}[h!]
    \centering
    \includegraphics[width=0.9\linewidth]{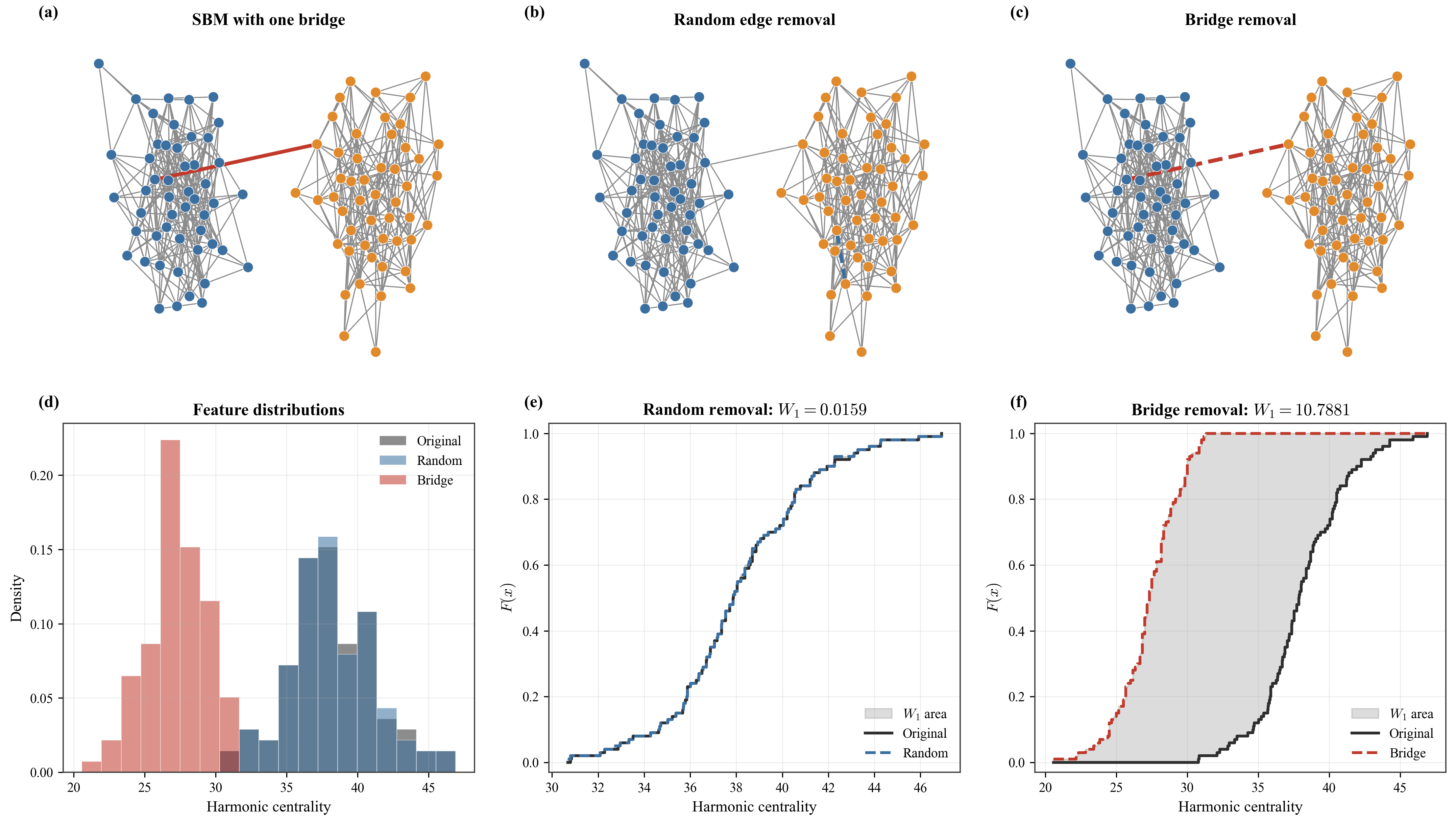}
    \caption{\textbf{(a)} The stochastic block model with two communities, intra-probability $p=0.18$, and one bridge (red solid). \textbf{(b), (c)} $G'$ after a random edge (blue-dashed) and the bridge (red-dashed) removal. \textbf{(d)} The histogram of harmonic centrality of $G$ and $G'$. \textbf{(e), (f)} The $W_1(\mu_G,\mu_{G'})$ is the area between the cumulative distribution functions of $\mu_G$ and $\mu_{G'}$ (gray colored). When an intra-community edge is removed, the cumulative distribution functions almost overlap, resulting in $W_1\approx 0.0159$ compared to $W_1\approx 10.7881$ in the bridge removal case.}
    \label{fig:placeholder}
\end{figure}

Unlike local, structural node features, Laplacian eigenvalues capture the total magnitude of an edge deletion but lose information about where the deletion occurs. The Wasserstein distance between graphs of the same size stays constant regardless of which edge is removed, making it invariant to the structural placement of the removal.

\begin{theorem}
    Let $G'=(V,E')$ be a subgraph of $G=(V,E)$, with $|V|=n$, and $|E'|=|E|-1$. If $\mu_G=\frac{1}{n}\sum_{i=1}^n \delta_{\lambda_i}$ and $\mu_{G'}=\frac{1}{n}\sum_{i=1}^n \delta_{\lambda_i'}$, then $W_1(\mu_G,\mu_{G'})=\frac{2}{n}$.
\end{theorem}
This theorem exposes a fundamental limitation of spectral empirical embeddings: every single edge removal produces the same $W_1$ distance, independent of the structural importance of the removed edge. The detailed proof of \textbf{Theorem 1} is provided in the supplementary material. Briefly, we utilize the interlacing property of the Laplacian spectrum (\cite{brouwer2011spectra}) and the closed form of $W_p(\mu_G,\mu_{G'})$ when the supports of $\mu_G$ and $\mu_{G'}$ have the same cardinality (\cite{peyre2019computational}).

\subsubsection{Metric measure space}\label{subsubsec3}

Instead of comparing distributions or graph spectra, we can embed a graph $G$ into a metric measure space (mm-space)
 $(V,d_V,\mu_G)$. Here, $d_V$ and $\mu_G$ are a metric and a measure in $V$, respectively. 

\begin{definition}[Gromov-Wasserstein Distance, \cite{memoli2011gromov}]
For $1\leq p<\infty$, the Gromov-Wasserstein distance between $(V,d_V,\mu_G)$ and $(V',d_{V'},\mu_{G'})$ is
\begin{align}
    GW_p(G,G')=\bigg(\min_{\pi\in \Pi(\mu_G,\mu_{G'})}\sum_{i,j=1}^n\sum_{k,l=1}^m |d_V(i,j)-d_{V'}(k,l)|^p\pi_{ik}\pi_{jl}\bigg)^\frac{1}{p}.
\end{align}
\end{definition}
The $GW_p$ is a metric in the set of all isomorphism classes of mm-spaces (\cite{memoli2011gromov}). The transport plan in $GW_p$ aligns two mm-spaces by minimizing internal distance distortion rather than the Euclidean costs. In contrast to $W_p$, $GW_p$ compares spaces with different dimensions or features (\textit{e.g.,} matching a 2D shape to a 3D shape). It is a nonconvex problem, and the closed form is available for $n=m$ and uniform weights in one dimension (\cite{vayer2019sliced}). Moreover, it generally lacks a closed form Monge map because the objective function is nonconvex (\cite{dumont2025existence}, \cite{memoli2024comparison}). In practice, the Sinkhorn algorithm adds a negative entropy term to approximate $GW_p$ by computing the entropic regularized distance (\cite{rioux2024entropic}).

\subsection{Gaussian Network Embedding}\label{subsec5}

\subsubsection{Graph signal distribution}\label{subsubsec4}

The $W_2$ between two Gaussians $\mathcal{N}(\mu_1,\Sigma_1)$ and $\mathcal{N}(\mu_2,\Sigma_2)$ has a closed form, and the Monge map is the affine transformation $T(x)=\mu_2+A(x-\mu_1)$, where $A=\Sigma_1^{-1/2}(\Sigma_1^{1/2}\Sigma_2\Sigma_1^{1/2})^{1/2}\Sigma_1^{-1/2}$ (\cite{peyre2019computational}). In particular, if the Gaussians have zero mean, then $W_2$ coincides with the Bures distance, a metric in the set of all positive semi-definite matrices (\cite{dowson1982frechet}, \cite{bhatia2019bures}). This closed form is widely used in graph signal processing (\cite{dong2016learning}) by embedding $G$ into a graph signal distribution $\mu=\mathcal{N}(0,L^\dagger)$, where $L^\dagger$ is the Moore-Penrose pseudoinverse.

\begin{definition}[Bures-Wasserstein Distance, \cite{petric2019got}]
The Bures-Wasserstein distance between graph signal distributions $\mu=\mathcal{N}(0,L^\dagger)$ and $\nu=\mathcal{N}(0,(L')^\dagger)$ is
\begin{align}
    BW(\mu,\nu)^2=\text{tr}\bigg[L^\dagger+(L')^\dagger-2\bigg((L^\dagger)^\frac{1}{2}(L')^\dagger(L^\dagger)^\frac{1}{2}\bigg)^\frac{1}{2}\bigg].
\end{align}
\end{definition}
In contrast to $GW_p$, $BW$ depends on the coordinate alignment used to represent the two Laplacian matrices. Consequently, when graph nodes lack a common labeling, $BW$ requires an explicit alignment or padding convention. Among the transport distances in this paper, $BW$ admits a closed-form expression in high dimensions.

Unlike the spectral distance $d_{\text{spec}}(G,G')=\sqrt{\sum_{i=1}^n (\lambda_i-(\lambda_i'))^2}$, $BW$ can measure the dissimilarity between some isospectral graphs. In \textbf{Fig. 2}, we plot the Laplacian spectra and the change of basis matrix of non-isomorphic isospectral graphs (\cite{vdam2003graphs}). Since $BW$ is a metric in the set of all positive semi-definite matrices, $BW(\mu,\nu)=0$ if and only if $L=L'$.

\vspace{-0.5cm}
\begin{figure}[h!]
    \centering
    \includegraphics[width=0.65\linewidth]{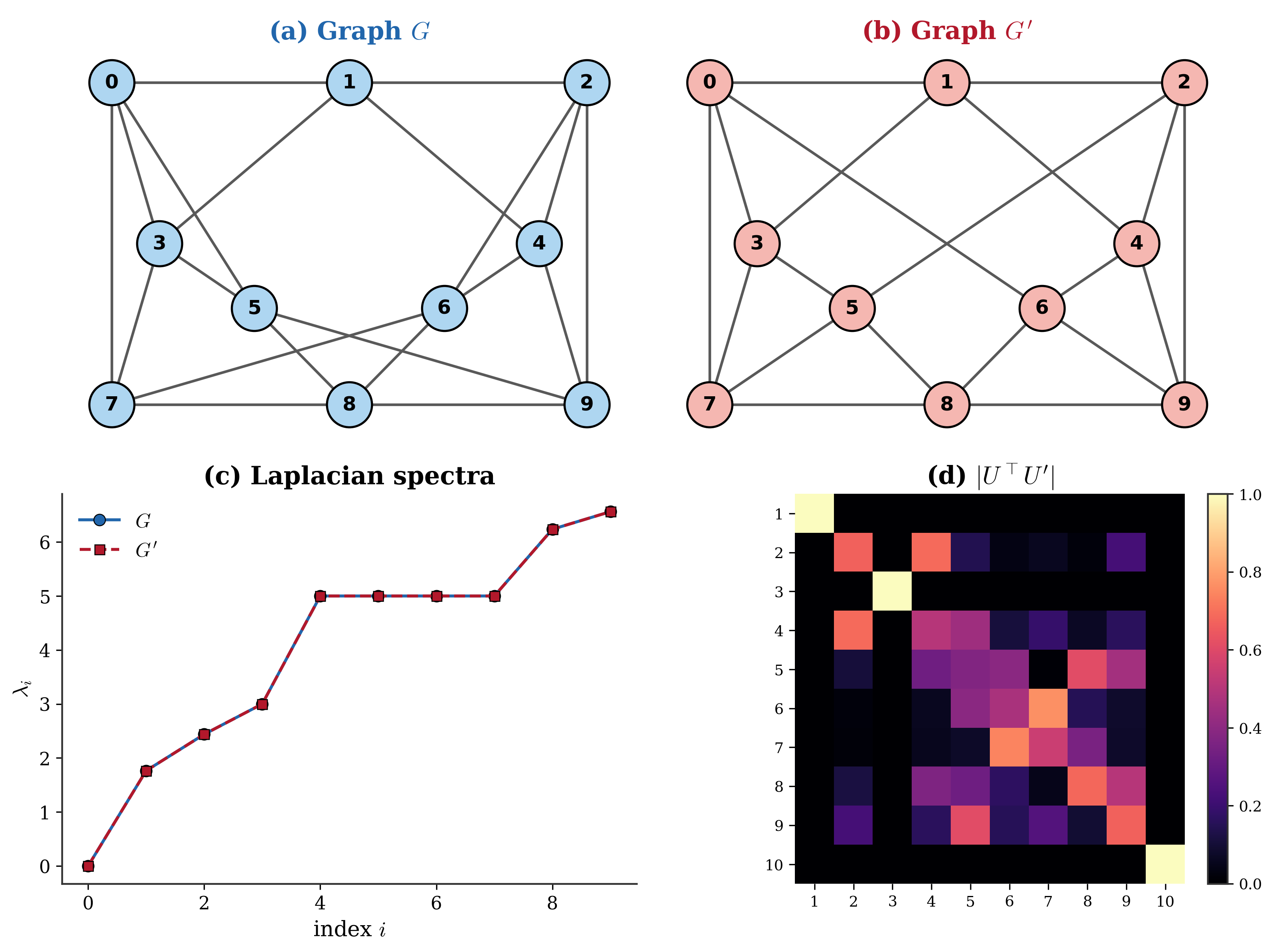}
    \caption{\textbf{(a), (b)} Example of nonisomorphic isospectral graphs $(G,G')$. \textbf{(c)} Same Laplacian spectrum of $L$ and $L'$: $\{0, 1.8, 2.4, 3, 5, 5, 5, 5, 6.2, 6.6\}$. \textbf{(d)} The heatmap of the change of basis matrix $|U^TU'|$. Since it is not the identity matrix, $U$ and $U'$ are not aligned. In this example, $d_{\textbf{spec}}(G,G')=0$ and $BW(\mu,\nu)\approx 0.24$.}
    \label{fig:placeholder}
\end{figure}

\subsubsection{Spectral bounds}\label{subsubsec5}

For connected graphs $G$ and $G'$, if $L$ and $L'$ share the same eigenvectors, then $BW(\mu,\nu)^2=\sum_{i=2}^n (1/\sqrt{\lambda_i}-1/\sqrt{\lambda_i'})^2$ (\cite{panaretos2020invitation}). The pseudoinverse of a Laplacian matrix inverts all non-zero eigenvalues and maps the zero eigenvalue to zero. In the following, we prove the bounds of $BW$ when $L$ and $L'$ do not have the same eigenvectors by the virtue of the matrix generalization of the \textit{Hardy-Littlewood-P\'olya rearrangement inequality} (\cite{yue2026matrix}).

\begin{theorem}
If $G=(V,E)$ and $G'=(V,E')$ with $|V|=n$ are connected, then
\begin{align*}
    \sum_{i=2}^n \bigg(\frac{1}{\sqrt{\lambda_i}}-\frac{1}{\sqrt{\lambda_i'}}\bigg)^2\leq BW(\mu,\nu)^2\leq \sum_{i=2}^n \bigg(\frac{1}{\sqrt{\lambda_{n-i+2}}}-\frac{1}{\sqrt{\lambda_i'}}\bigg)^2.
\end{align*}
\end{theorem}
These spectral bounds do not require computing the eigenvectors of the Laplacian matrices, in contrast to (4) that requires full spectral decompositions. 

In most cases, the Laplacian eigenvectors are not aligned under graph perturbation. We prove that the spectral lower bound in \textbf{Theorem 2} is tight whenever the Laplacian eigenbases are close under the spectral norm by \textit{Weyl's inequality}.

\begin{theorem}
Let $G=(V,E)$ and $G'=(V,E')$ with $|V|=n$ be connected. If $U^TU'=I+E$, where $I\in \mathbb{R}^{n\times n}$ is the identity matrix and $E$ satisfies $\|E\|_2=\epsilon$ ($\|\cdot\|_2$ spectral norm), then
\begin{align*}
    BW(\mu,\nu)^2-\sum_{i=2}^n \bigg(\frac{1}{\sqrt{\lambda_i}}-\frac{1}{\sqrt{\lambda_i'}}\bigg)^2\leq \frac{2(n-1)\sqrt{2\epsilon+\epsilon^2}}{\sqrt{\lambda_{2}\lambda_2'}}.
\end{align*}
\end{theorem}
Because eigenvectors are defined only up to sign and, for repeated eigenvalues, rotations within eigenspaces, the condition $U^TU'=I+E$ is understood after choosing compatible eigenbases. The detailed proofs of \textbf{Theorem 2} and \textbf{Theorem 3} are provided in the supplementary material.

\section{Numerical Experiments}\label{sec4}

\subsubsection{Circulant graphs}\label{subsubsec6}

We find an example of a graph when the spectral lower bound in \textbf{Theorem 2} is tight. A circulant graph can be transformed into other members of its graph class by appropriately modifying its edge sets. Every circulant graph $C_n(1,\cdots,m)$ with $n$ nodes has the same Laplacian eigenvectors regardless of the number of jumps $m$. The Laplacian matrix is diagonalized by real Fourier basis vectors, and the eigenvalues are $\lambda_j=2\sum_{k=1}^m (1-\cos\frac{2\pi kj}{n})$ for all $1\leq j\leq n$ (\cite{kotzagiannidis2019splines}). If $\mu$ and $\nu$ are the graph signal distributions of $G=C_n(1,\cdots, m)$ and $G'=C_n(1,\cdots, m')$, then $BW(\mu,\nu)^2$ has the closed-form expression by \textbf{Theorem 2}.
\begin{align}
    \sum_{j=1}^{n-1} \bigg[\bigg(\sqrt{2\sum_{k=1}^m\bigg(1-\cos\frac{2\pi kj}{n}\bigg)}\bigg)^{-1}-\bigg(\sqrt{2\sum_{k=1}^{m'} \bigg(1-\cos\frac{2\pi kj}{n}\bigg)}\bigg)^{-1}\bigg]^2.
\end{align}
To validate the tightness of the spectral lower bound in \textbf{Theorem 2}, we compute the Bures-Wasserstein distance between $C_{16}(1,2)$ and $C_{16}(1,\cdots,m')$ and observe how the distance changes as the jump size $m'$ grows in \textbf{Fig. 3}.

\begin{figure}[h!]
\begin{minipage}[c]{0.5\linewidth}
\includegraphics[width=\linewidth]{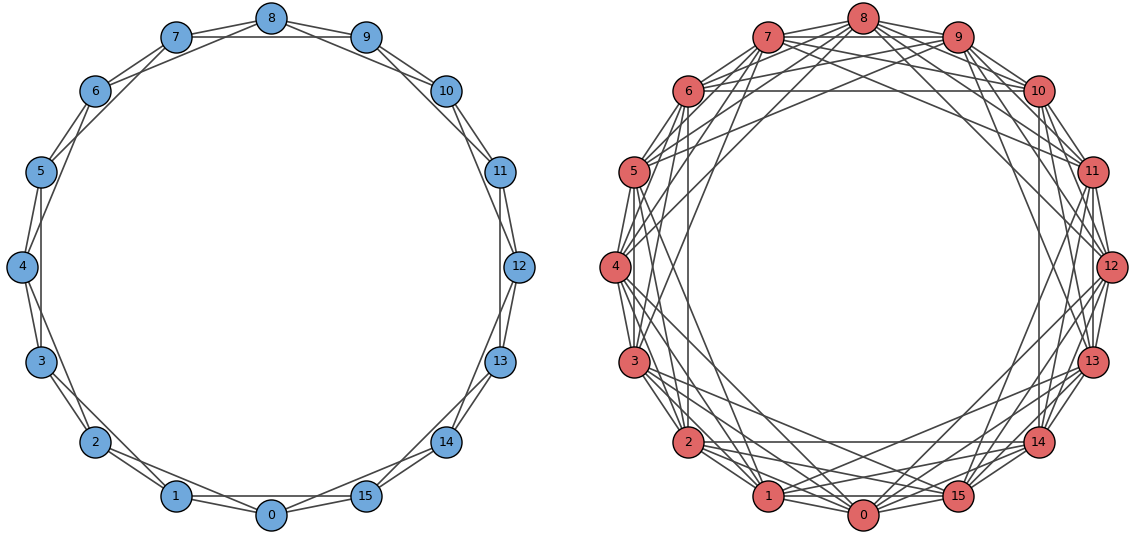}
%\caption{Two circulant graphs with 16 nodes}
\end{minipage}
\hfill
\begin{minipage}[c]{0.5\linewidth}
\includegraphics[width=\linewidth]{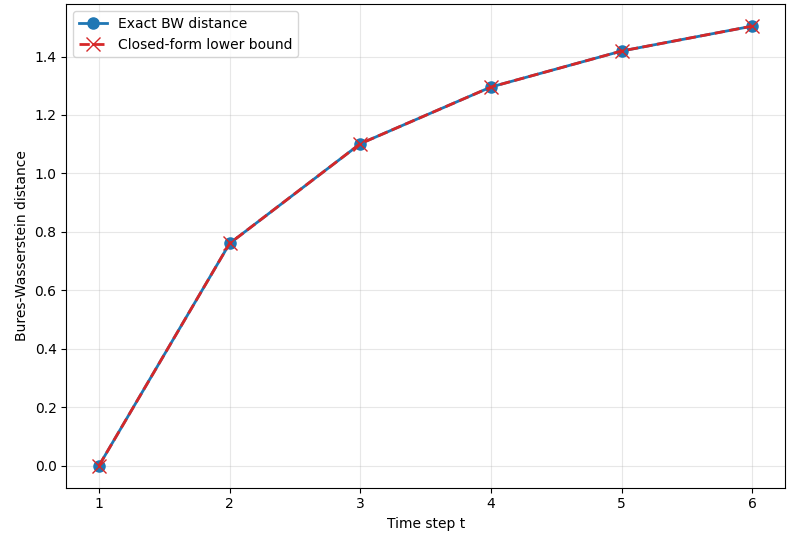}
%\caption{$BW^2$ and its lower bound}
\end{minipage}%
\caption{\textbf{(left)} Circulant graphs $C_{16}(1,2)$ and $C_{16}(1,2,3,4)$. \textbf{(right)} The Bures-Wasserstein distance and the spectral lower bound in \textbf{Theorem 2} between $C_{16}(1,2)$ and $C_{16}(1,\cdots,m')$ for $m'$ ranging between 2 and 7.}
\end{figure}

\subsubsection{Network clustering}\label{subsubsec7}

We cluster networks from the SYNTHIE dataset (\cite{morris2016faster}) by transport distances and compare their clustering performance against the graph edit distance, approximated via bipartite matching (\cite{riesen2009approximate}). We choose $K$-medoids instead of the traditional $K$-means because the mapping from the set of all graphs to the set of all probability measures is non-injective (\textit{e.g.,} isospectral graphs in \textbf{Fig. 2}). The dataset has 4 class labels for $400$ networks with 90 or 100 nodes each and different edge densities. 

For $W_1$, we choose the following local, global, and attribute node features: harmonic centrality ($W_{1,h}$), Laplacian eigenvalues ($W_{1,L}$), and the 15-dimensional node attributes from the dataset ($W_{1,a}$). For $BW$, we zero-pad $L^\dagger$ to keep dimensions consistent and label the nodes sequentially since the dataset does not contain node labels. Finally, $GW_2$ adopts a diffusion distance (\cite{xiao2005characterising}) as the ground metric in (3). The distances $W_{1,a}$ and $GW_2$ lacking a closed-form expression were implemented using the POT library (\cite{flamary2026pot}). We compute the $400\times 400$ pairwise distance matrix per distance and use a greedy search method, PAM, to cluster the graphs into $K=4$ groups (\cite{kaufman}) in \textbf{Fig. 4}. We tabulated the classification score against the true labels with ARI, AMI, and V-measure in \textbf{Table 1}.

As shown in \textbf{Fig. 4}, in contrast to the approximate edit distance and $W_{1,a}$, these four distances $W_{1,h}$, $W_{1,L}$, $BW$, and $GW_2$ successfully measured graph dissimilarity and classified the graphs based on their topology. The clustering experiment suggests that, for this dataset, coarse topological structure dominates the particular choice of transport distance, whereas the node attributes supplied by SYNTHIE do not improve clustering performance.

\begin{table}[h!]
\centering
\caption{4-medoids clustering performance on SYNTHIE by distance}\label{tab1}%
\begin{tabular}{@{}|c|c|c|c|c|c|c|@{}}
\toprule
Clustering Score & Edit distance & $W_{1,h}$(local) & $W_{1,L}$(global) & $W_{1,a}$(attribute) & $BW$ & $GW_2$ \\
\midrule
ARI & -0.001 & 0.333 & 0.334 & 0.197 & 0.334 & 0.334 \\
AMI & 0.000 & 0.498 & 0.498 & 0.237 & 0.499 & 0.499 \\
V-measure & 0.009 & 0.502 & 0.503 & 0.243 & 0.503 & 0.503 \\
\bottomrule
\end{tabular}
\end{table}

\begin{figure}[h!]
    \centering
    \includegraphics[width=1.0\linewidth]{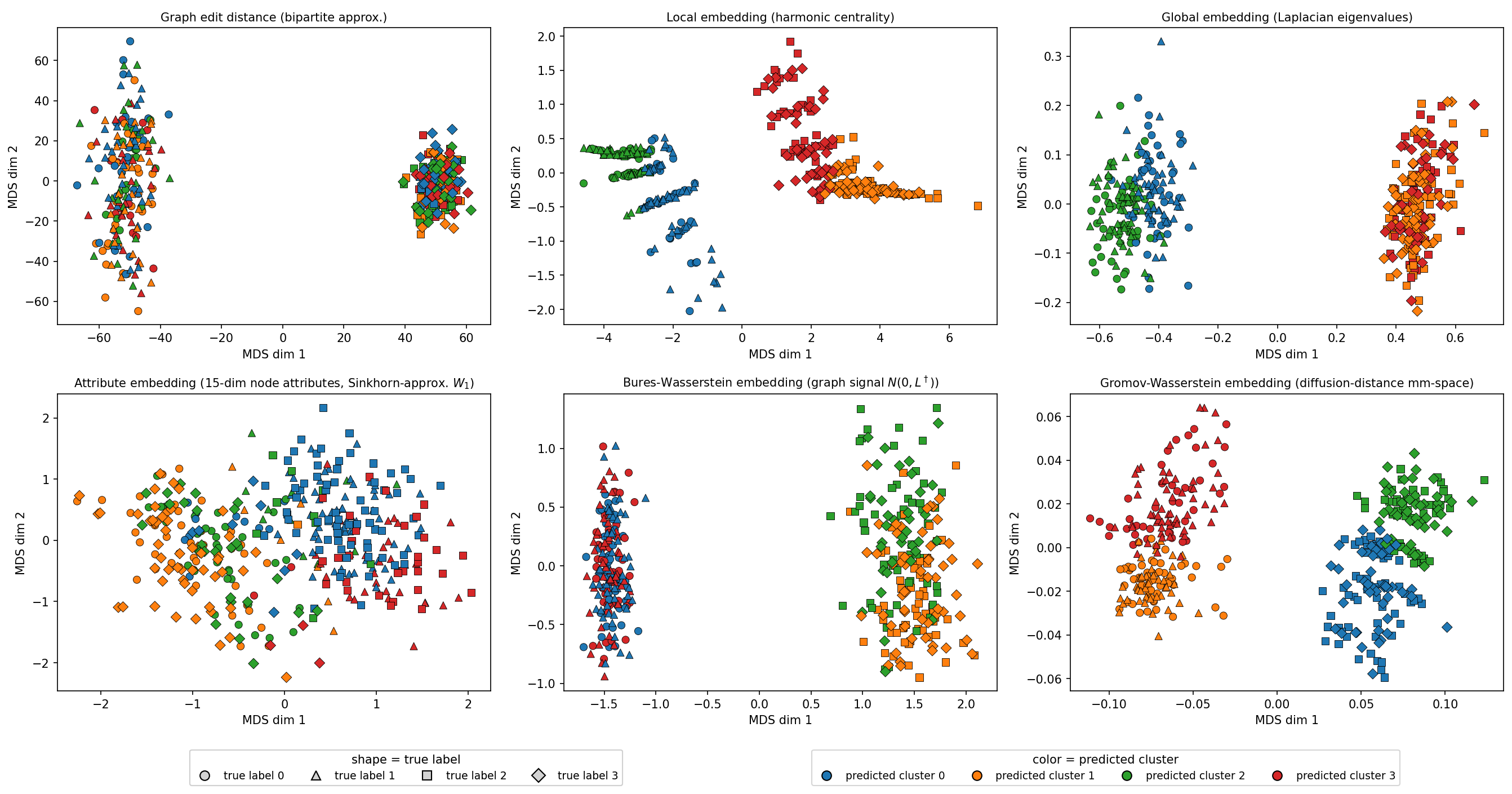}
    \caption{2D multidimensional scaling plot of SYNTHIE graphs colored by 4-medoids clustering. From left to right, the first row displays clustering for edit distance, $W_{1,h}$, and $W_{1,L}$. The second row displays clustering for $W_{1,a}$, $BW$, and $GW_2$.}
    \label{fig:placeholder}
\end{figure}

\subsubsection{Temporal network}\label{subsubsec8}

The Enron corpus (\cite{klimt2004introducing}) contains weekly email networks from roughly 150 senior executives of the Enron corporation. We adapt the numerical experiment from \cite{monnig2018resistance} that computes the resistance perturbation distance, DeltaCon, and $d_{\text{CAD}}$ between two consecutive weekly email networks. While the original experiment forces a fixed node count across all weeks -- padding inactive weeks with isolated nodes -- our version builds each week's adjacency matrix using only active senders and receivers for the Wasserstein distance $W_1$ and Gromov-Wasserstein distance $GW_2$. 

As in the clustering experiment, we compare the graphs using the same distances. The only difference is that (2) is computed between degree distributions. The resulting normalized distances between consecutive weekly email networks are plotted in \textbf{Fig. 5}. We observe that when no labeled event occurs, $W_1$ spiked while $GW_2$ remains flat across the entire dataset. In contrast, the larger changes between consecutive emails detected by $BW$ are predictive that lead to the company's ultimate collapse. 

In addition, we plot the transport plans of $W_1$ and $GW_2$ during the week when Dynegy agrees to buy Enron in \textbf{Fig. 6}. These transport plans identify which nodes in the previous week shift a huge amount of mass to nodes in the current week, visualizing pinpoint structural changes. The Bures-Wasserstein distance lacks a node-to-node transport plan because it restricts comparisons to Gaussian distributions, operating on covariance matrices rather than empirical measures or individual nodes in a graph. 

\begin{figure}[h!]
    \centering
    \includegraphics[width=1.0\linewidth]{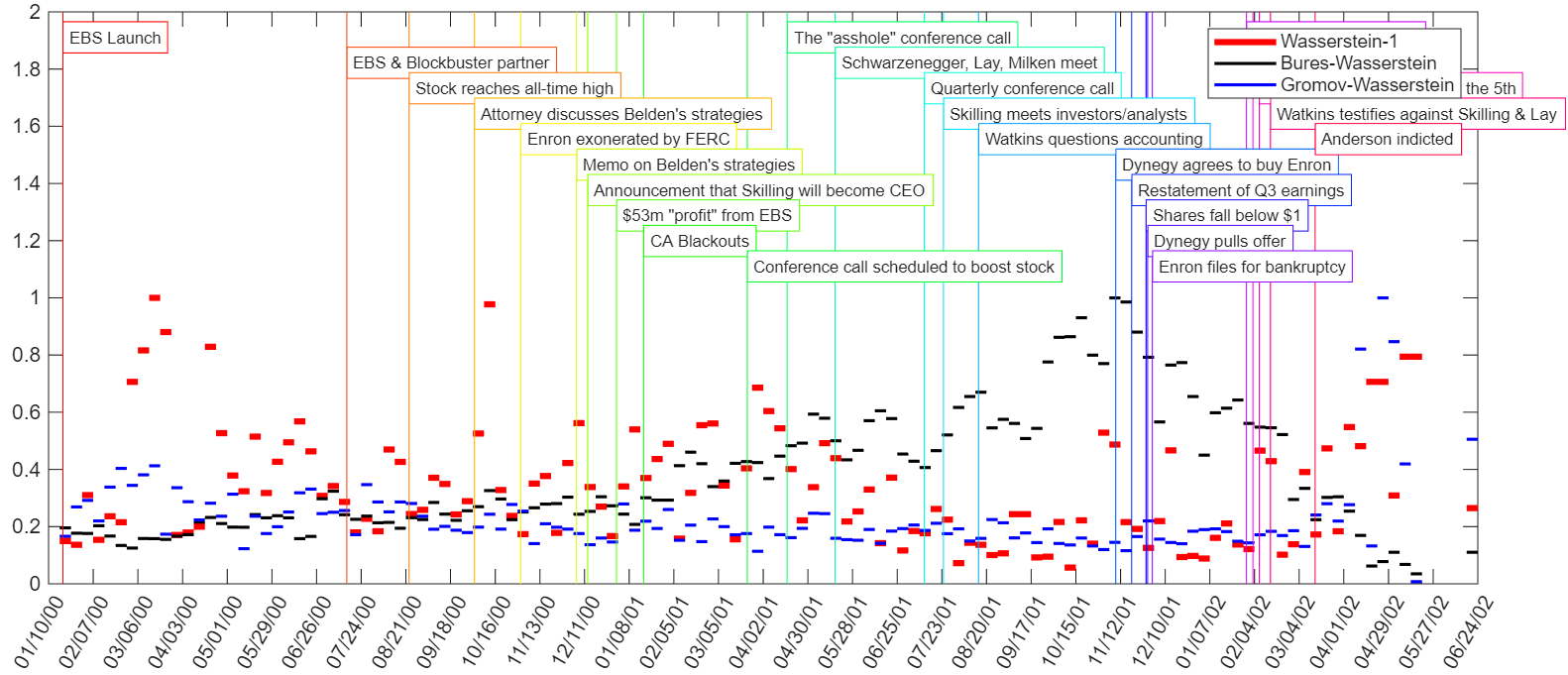}
    \caption{Normalized distances $W_1$ (red), $BW$ (black), and $GW_2$ (blue) between consecutive Enron email networks from 2000 to 2002.}
    \label{fig:placeholder}
\end{figure}

\begin{figure}[h!]
\begin{minipage}[c]{0.4\linewidth}
\includegraphics[width=\linewidth]{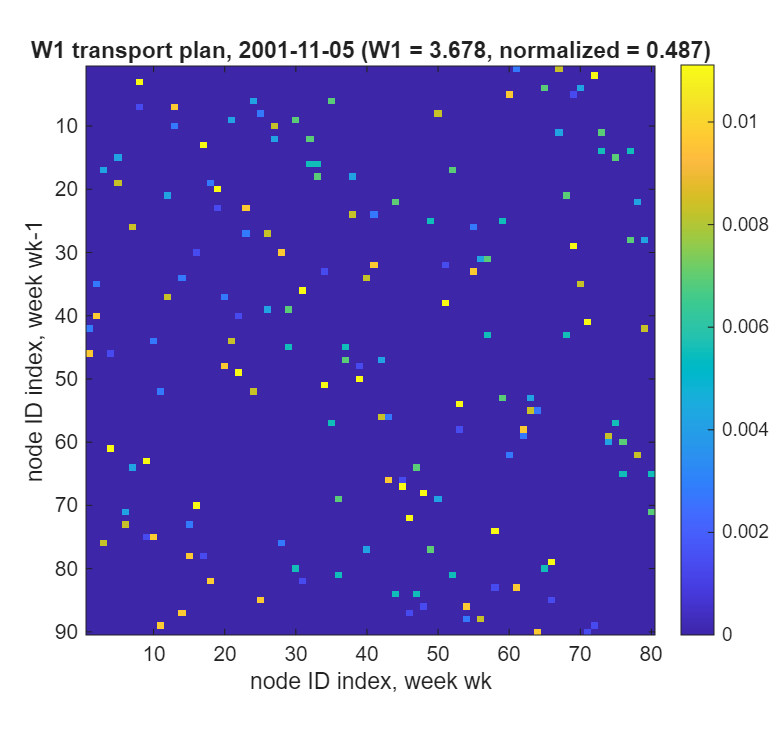}
%\caption{}
\end{minipage}
\hfill
\begin{minipage}[c]{0.37\linewidth}
\includegraphics[width=\linewidth]{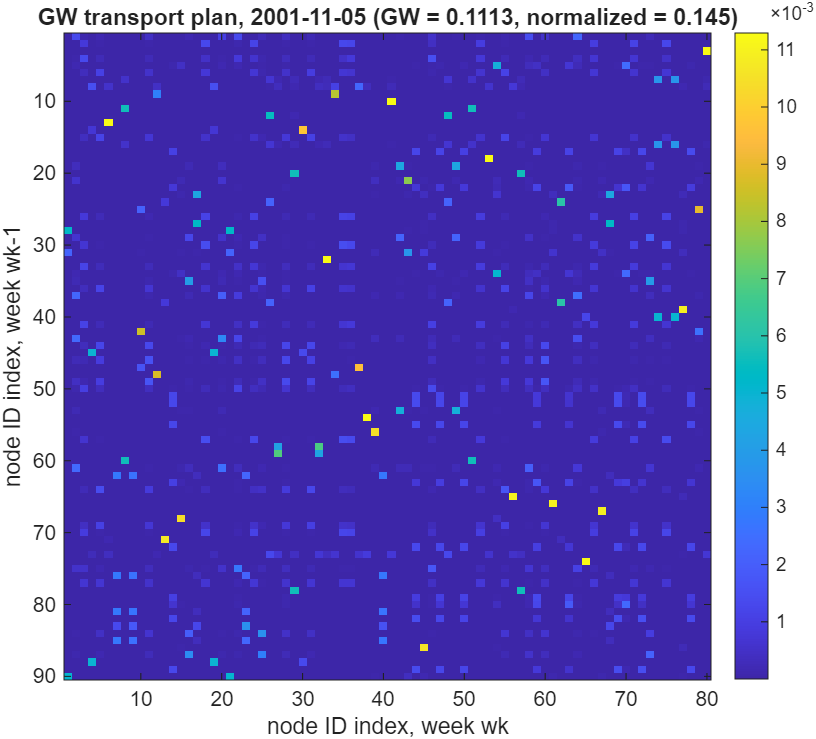}
%\caption{}
\end{minipage}%
\caption{Transport plans of $W_1$ and $GW_2$ during the week of 2001/11/05 with event label: Dynegy agrees to buy Enron. In that week, $W_1\approx 0.487$, $BW=1$, and $GW\approx 0.145$.}
\end{figure}

\section{Discussion and Conclusion}\label{sec5} 

We outline several directions for future work. Although our study examines the one-dimensional closed form and the Sinkhorn algorithm, computing these distances for large networks in high dimensions remains a key challenge. Moreover, constructing high-dimensional network embeddings that capture both node attributes and graph topology via optimal transport remains an open field (see \cite{grover2016node2vec}, \cite{belkin2003laplacian}, and \cite{vayer2020fused} for more details).

Beyond these challenges, a unique advantage of transport distances is their ability to not only quantify structural differences between graphs but also reveals why networks differ via the transport plan. We summarize the characteristics of the transport distances in \textbf{Table 2}. Altogether, transport distances allow researchers to compare networks of different sizes via closed-form expressions and visualize network dissimilarity.

\begin{table}[h!]
\centering
\caption{Summary of Transport Distances for Network Comparison}\label{tab1}%
\begin{tabular}{@{}|c|c|c|c|@{}}
\toprule
Property & $W_1$ & $BW$ & $GW_2$ \\
\midrule
Embedding & Feature distribution & Pseudoinverse & Metric measure space \\
$n\neq m$ & \checkmark & Requires padding & \checkmark \\
Permutation invariant & \checkmark & alignment-dependent & \checkmark \\
Closed form & 1D/Gaussian & \checkmark & 1D, uniform weights, $n=m$ \\
Transport plan & \checkmark & \ding{55} & \checkmark \\
\bottomrule
\end{tabular}
\end{table}


\bibliographystyle{spmpsci} % We choose the "plain" reference style
\bibliography{refs} % Entries are in the refs.bib file

\newpage
\appendix
\setcounter{proposition}{0}
\renewcommand{\theproposition}{\arabic{proposition}}

\setcounter{theorem}{0}
\renewcommand{\thetheorem}{\arabic{theorem}}

\section{Supplementary Material}

\subsection{Proof of Theorem 1}

\begin{theorem}
    Let $G'=(V,E')$ be a subgraph of $G=(V,E)$, with $|V|=n$, and $|E'|=|E|-1$. If $\mu_G=\frac{1}{n}\sum_{i=1}^n \delta_{\lambda_i}$ and $\mu_{G'}=\frac{1}{n}\sum_{i=1}^n \delta_{\lambda_i'}$, then
    \begin{align*}
        W_1(\mu_G,\mu_{G'})=\frac{2}{n}.
    \end{align*}
\end{theorem}

\begin{proof}
Let $L$ and $L'$ be the Laplacian matrices of $G$ and $G'$, respectively. Since only one edge, say $(i,j)\in E$, is removed, the perturbed Laplacian matrix is written as
\begin{align*}
    L'=L-(e_i-e_j)(e_i-e_j)^T.
\end{align*}
Here, $e_i$ is the $i$th standard basis vector, and the outer product $(e_i-e_j)(e_i-e_j)^T$ is a rank-1 matrix with eigenvalues 2 and 0 with respective algebraic multiplicity of $1$ and $n-1$. The eigenvalues of $L'$ interleave with the eigenvalues of $L$ (\textbf{Proposition 3.2.1} in \cite{brouwer2011spectra}):
\begin{align*}
    \lambda_1'\leq \lambda_1\leq \lambda_2'\leq \lambda_2\leq\cdots\leq \lambda_n'\leq \lambda_n.
\end{align*}
For each $i$, $\lambda_i-\lambda_i'\geq 0$. Using the closed form (\textbf{Remark 2.28} in \cite{peyre2019computational}),
\begin{align*}
    W_1(\mu_G,\mu_{G'})=\frac{1}{n}\sum_{i=1}^n (\lambda_i-\lambda_i')=\frac{1}{n}\text{tr}(L-L')=\frac{2}{n}.
\end{align*}
Hence, it is proved.
\end{proof}

\newpage

\subsection{Proof of Theorem 2}

\begin{theorem}
    If $G=(V,E)$ and $G'=(V,E')$ with $|V|=n$ are connected, then
    \begin{align*}
        \sum_{i=2}^n \bigg(\frac{1}{\sqrt{\lambda_i}}-\frac{1}{\sqrt{\lambda_i'}}\bigg)^2\leq BW(\mu,\nu)^2\leq \sum_{i=2}^n \bigg(\frac{1}{\sqrt{\lambda_{n-i+2}}}-\frac{1}{\sqrt{\lambda_i'}}\bigg)^2.
    \end{align*}
\end{theorem}

\begin{proof}
Let $M=(L^\dagger)^\frac{1}{2}(L')^\dagger(L^\dagger)^\frac{1}{2}$. Taking the Moore-Penrose pseudoinverse and square root of the spectral decompositions $L=U\Lambda U^T$ and $L'=U'\Lambda'(U')^T$,
\begin{align*}
    (L^\dagger)^\frac{1}{2}=U(\Lambda^\dagger)^{\frac{1}{2}}U^T,\quad (L')^\dagger=U'(\Lambda')^\dagger(U')^T,
\end{align*}
where $\Lambda^\dagger=\text{diag}(0,\lambda_2^{-1},\cdots,\lambda_n^{-1})$ and $(\Lambda')^\dagger=\text{diag}(0,(\lambda_2')^{-1},\cdots,(\lambda_n')^{-1})$. Set $O\coloneq U^TU'$ and rewrite $M$.
\begin{align*}
    M=U(\Lambda^\dagger)^\frac{1}{2}O(\Lambda')^\dagger O^T(\Lambda^\dagger)^\frac{1}{2}U^T.
\end{align*}
Since eigenvalues are invariant under an orthogonal transformation,
\begin{align*}
    \forall i\in \llbracket n \rrbracket,\quad \lambda_i(M)=\lambda_i((\Lambda^\dagger)^\frac{1}{2}O(\Lambda')^\dagger O^T(\Lambda^\dagger)^\frac{1}{2}).
\end{align*}
We invoke the matrix generalization of the \textit{Hardy-Littlewood-P\'olya rearrangement inequality} (\textbf{Theorem 3} in \cite{yue2026matrix}). Define $f(s)=\sqrt{s}$ for $s>0$. Then $f$ is differentiable in $(0,\infty)$, $sf'(s)=\frac{\sqrt{s}}{2}$ is increasing, and $f$ is right-continuous at $0$. Hence, for positive semi-definite matrices $A=O(\Lambda')^\dagger O^T$ and $B=\Lambda^\dagger$,
\begin{align*}
    \sum_{i=1}^n f(\lambda_i(A)\lambda_{n-i+1}(B))\leq \sum_{i=1}^n f(\lambda_i(B^\frac{1}{2}AB^\frac{1}{2}))\leq \sum_{i=1}^n f(\lambda_i(A)\lambda_i(B)).
\end{align*}
Since $\lambda_1(A)=\lambda_1(B)=0$, $\lambda_i(A)=(\lambda_i')^{-1}$, $\lambda_i(B)=(\lambda_i)^{-1}$, and $\lambda_i(B^\frac{1}{2}AB^\frac{1}{2})=\lambda_i(M)$,
\begin{align*}
    \sum_{i=2}^n \sqrt{\frac{1}{\lambda_i'\lambda_{n-i+2}}}\leq \sum_{i=1}^n \sqrt{\lambda_i(M)}\leq \sum_{i=2}^n \sqrt{\frac{1}{\lambda_i'\lambda_i}}.
\end{align*}
By \textit{Spectral Mapping Theorem}, $\text{tr}(M^\frac{1}{2})=\sum_{i=1}^n \sqrt{\lambda_i(M)}$. Now, we substitute the chain of inequalities in the Bures-Wasserstein distance. For the lower bound,
\begin{align*}
    BW(\mu,\nu)^2\geq \sum_{i=2}^n \frac{1}{\lambda_i}+\sum_{i=2}^n \frac{1}{\lambda_i'}-2\sum_{i=2}^n \sqrt{\frac{1}{\lambda_i'\lambda_i}}=\sum_{i=2}^n \bigg(\sqrt{\frac{1}{\lambda_i}}-\sqrt{\frac{1}{\lambda_i'}}\bigg)^2.
\end{align*}
Similarly for the upper bound,
\begin{align*}
    BW(\mu,\nu)^2&\leq \sum_{i=2}^n \frac{1}{\lambda_i}+\sum_{i=2}^n \frac{1}{\lambda_i'}-2\sum_{i=2}^n \sqrt{\frac{1}{\lambda_i'\lambda_{n-i+2}}}\\
    &=\sum_{i=2}^n \frac{1}{\lambda_{n-i+2}}+\sum_{i=2}^n \frac{1}{\lambda_i'}-2\sum_{i=2}^n \sqrt{\frac{1}{\lambda_i'\lambda_{n-i+2}}}\\
    &=\sum_{i=2}^n \bigg(\sqrt{\frac{1}{\lambda_{n-i+2}}}-\sqrt{\frac{1}{\lambda_i'}}\bigg)^2.
\end{align*}
Therefore, it is proved.
\end{proof}

\newpage
\subsection{Proof of Theorem 3}

\begin{theorem}
Let $G=(V,E)$ and $G'=(V,E')$ with $|V|=n$ be connected. If $U^TU'=I+E$, where $I\in \mathbb{R}^{n\times n}$ is the identity matrix and $E$ satisfies $\|E\|_2=\epsilon$ ($\|\cdot\|_2$ spectral norm), then
\begin{align*}
    BW(\mu,\nu)^2-\sum_{i=2}^n \bigg(\frac{1}{\sqrt{\lambda_i}}-\frac{1}{\sqrt{\lambda_i'}}\bigg)^2\leq \frac{2(n-1)\sqrt{2\epsilon+\epsilon^2}}{\sqrt{\lambda_{2}\lambda_2'}}.
\end{align*}
\end{theorem}

\begin{proof}
Let $M=(L^\dagger)^\frac{1}{2}(L')^\dagger(L^\dagger)^\frac{1}{2}$. Subtracting the lower bound in \textbf{Theorem 2} from the squared Bures-Wasserstein distance,
\begin{align*}
    BW(\mu,\nu)^2-\sum_{i=2}^n\bigg(\frac{1}{\sqrt{\lambda_i}}-\frac{1}{\sqrt{\lambda_i'}}\bigg)^2&=2\sum_{i=2}^n \frac{1}{\sqrt{\lambda_i\lambda_i'}}-2\text{tr}\bigg[(L^\frac{\dagger}{2}(L')^\dagger L^\frac{\dagger}{2})^\frac{1}{2}\bigg]\\
    &=2\sum_{i=2}^n \bigg(\frac{1}{\sqrt{\lambda_i\lambda_i'}}-\sqrt{\lambda_i(M)}\bigg).
\end{align*}
Here, $\lambda_i(M)=\lambda_i((\Lambda^\dagger)^\frac{1}{2}O(\Lambda')^{\dagger}O^T(\Lambda^\dagger)^\frac{1}{2})$, where we set $O\coloneq U^TU'$, $\Lambda^\dagger=\text{diag}(0,\lambda_2^{-1},\cdots,\lambda_n^{-1})$, and $(\Lambda')^\dagger=\text{diag}(0,(\lambda_2')^{-1},\cdots,(\lambda_n')^{-1})$. Inserting $O=I+E$ in $(\Lambda^\dagger)^\frac{1}{2}O(\Lambda')^{\dagger}O^T(\Lambda^\dagger)^\frac{1}{2}$,
\begin{align*}
    &(\Lambda^\dagger)^\frac{1}{2}(I+E)(\Lambda')^{\dagger}(I+E)^T(\Lambda^\dagger)^\frac{1}{2}=\bigg[(\Lambda^\dagger)^\frac{1}{2}(\Lambda')^{\dagger}+(\Lambda^\dagger)^\frac{1}{2}E(\Lambda')^{\dagger}\bigg](I+E^T)(\Lambda^\dagger)^\frac{1}{2}\\
    &=\Lambda^{-1}(\Lambda')^{\dagger}+(\Lambda^\dagger)^\frac{1}{2}(\Lambda')^{\dagger}E^T(\Lambda^\dagger)^\frac{1}{2}+(\Lambda^\dagger)^\frac{1}{2}E(\Lambda')^{\dagger}(\Lambda^\dagger)^\frac{1}{2}+(\Lambda^\dagger)^\frac{1}{2}E(\Lambda')^{\dagger}E^T(\Lambda^\dagger)^\frac{1}{2}.
\end{align*}
Write the diagonal matrix $D\coloneq \Lambda^\dagger(\Lambda')^{\dagger}$ and 
\begin{align*}
    \Delta=(\Lambda^\dagger)^\frac{1}{2}(\Lambda')^{\dagger}E^T(\Lambda^\dagger)^\frac{1}{2}+(\Lambda^\dagger)^\frac{1}{2}E(\Lambda')^{\dagger}(\Lambda^\dagger)^\frac{1}{2}+(\Lambda^\dagger)^\frac{1}{2}E(\Lambda')^{\dagger}E^T(\Lambda^\dagger)^\frac{1}{2}.
\end{align*}
By triangle inequality and the sub-multiplicative property of the spectral norm,
\begin{align*}
    \|\Delta\|_2&\leq \|(\Lambda^\dagger)^\frac{1}{2}(\Lambda')^{\dagger}E^T(\Lambda^\dagger)^\frac{1}{2}\|_2+\|(\Lambda^\dagger)^\frac{1}{2}E(\Lambda')^{\dagger}(\Lambda^\dagger)^\frac{1}{2}\|_2+\|(\Lambda^\dagger)^\frac{1}{2}E(\Lambda')^{\dagger}E^T(\Lambda^\dagger)^\frac{1}{2}\|_2\\
    &\leq (2\epsilon+\epsilon^2) \|\Lambda^{\dagger}\|_2\|(\Lambda')^{\dagger}\|_2\\
    &=(2\epsilon+\epsilon^2)\sqrt{\lambda_{\text{max}}((\Lambda^{\dagger})^T\Lambda^{\dagger})\lambda_{\text{max}}(((\Lambda')^{\dagger})^T(\Lambda')^{\dagger})}\\
    &=\frac{2\epsilon+\epsilon^2}{\lambda_2\lambda_2'}.
\end{align*}
Set $\lambda_i(D)=1/(\lambda_{n-i+2}\lambda_{n-i+2}')$ so that both the non-zero eigenvalues of $D$ and $M$ are in ascending order. By \textit{Weyl's inequality,}
\begin{align*}
    |\lambda_i(M)-\lambda_i(D)|=\bigg|\lambda_i(M)-\frac{1}{\lambda_{n-i+2}\lambda_{n-i+2}'}\bigg|\leq \|\Delta\|_2\leq \frac{2\epsilon+\epsilon^2}{\lambda_2\lambda_2'}.
\end{align*}
Taking the square root,
\begin{align*}
    \bigg|\sqrt{\lambda_i(M)}-\frac{1}{\sqrt{\lambda_{n-i+2}\lambda_{n-i+2}'}}\bigg|\leq \sqrt{\bigg|\lambda_i(M)-\frac{1}{\lambda_{n-i+2}\lambda_{n-i+2}'}\bigg|}\leq \sqrt{\frac{2\epsilon+\epsilon^2}{\lambda_2\lambda_2'}}.
\end{align*}
Multiplying both sides by 2 and taking the sum,
\begin{align*}
    2\sum_{i=2}^n \bigg(\frac{1}{\sqrt{\lambda_i\lambda_i'}}-\sqrt{\lambda_i(M)}\bigg)\leq 2\sum_{i=2}^n\bigg|\sqrt{\lambda_i(M)}-\frac{1}{\sqrt{\lambda_{n-i+2}\lambda_{n-i+2}'}}\bigg|\leq 2\sum_{i=2}^n \sqrt{\frac{2\epsilon+\epsilon^2}{\lambda_2\lambda_2'}}.
\end{align*}
Therefore, it is proved.
\end{proof}

\end{document}